\documentclass{article}

 \usepackage[dblblindworkshop, final]{neurips_2025}

\usepackage[utf8]{inputenc} %
\usepackage[T1]{fontenc}    %
\usepackage{hyperref}       %
\usepackage{url}            %
\usepackage{booktabs}       %
\usepackage{amsfonts}       %
\usepackage{nicefrac}       %
\usepackage{microtype}      %
\usepackage{xcolor}         %
\usepackage{graphicx}  %
\usepackage[dvipsnames]{xcolor}  %
\usepackage{array}  %
\usepackage{graphicx}  %
\usepackage{float}  %
\usepackage{subfig}  %
\usepackage{amsmath}  %
\usepackage{amsthm}   %
\usepackage{amsfonts} %
\usepackage{calc}  %
\usepackage{bm}  %
\usepackage{tikz}
\usetikzlibrary{positioning}  %
\usepackage{booktabs}

\usepackage{threeparttable}
\usepackage{adjustbox}
\newcommand{\unif}{\text{Unif}} 
\newcommand{\unifpm}{\text{Unif}_{\pm}} 
\newcommand{\Ncal}{\mathcal{N}}
\newcommand{\xb}{\bm{x}}
\newcommand{\wb}{\bm{w}}

\workshoptitle{AI for Tabular Data}
\title{Does TabPFN Understand Causal Structures?}

\author{%
    Omar Swelam$^1$ \quad
    Lennart Purucker$^1$\quad
    Jake Robertson$^{3,2,1}$ \quad
Hanne Raum$^{1}$ \\
\textbf{Joschka Boedecker}$^1$ \quad
\textbf{Frank Hutter}$^{2,3,1}$
\\
$^1$University of Freiburg \quad
$^2$Prior Labs \quad
$^3$ELLIS Institute Tübingen 
\\
 $\texttt{swelamo@informatik.uni-freiburg.de}$\\
}

\begin{document}

\maketitle

\begin{abstract}
    Causal discovery is fundamental for multiple scientific domains, yet extracting causal information from real world data remains a significant challenge. Given the recent success on real data, we investigate whether TabPFN, a transformer-based tabular foundation model pre-trained on synthetic datasets generated from structural causal models, encodes causal information in its internal representations. We develop an adapter framework using a learnable decoder and causal tokens that extract causal signals from TabPFN's frozen embeddings and decode them into adjacency matrices for causal discovery. Our evaluations demonstrate that TabPFN's embeddings contain causal information, outperforming several traditional causal discovery algorithms, with such causal information being concentrated in mid-range layers. These findings establish a new direction for interpretable and adaptable foundation models and demonstrate the potential for leveraging pre-trained tabular models for causal discovery.
\end{abstract}

\section{Introduction}
    \label{introduction}
   
    Traditional causal discovery methods face substantial challenges and impose assumptions that are often unverifiable in practice \citep{spirtes2000pc, chickering2002ges}. Meanwhile, tabular foundation models (TFMs), such as TabPFN \citep{hollmann2022tabpfn, hollmann2025tabpfnv2}, have shown remarkable performance and generalization on tabular data tasks despite having been pre-trained only on synthetic data. Which raises a question: do the internal representations of TabPFN encode causal knowledge beyond statistical correlations?
    We investigate whether TabPFN's (specifically TabPFNv2 \citep{ hollmann2025tabpfnv2}) representations contain causal information by developing a framework that learns to extract this information for causal discovery. Our framework employs learnable dual-attention decoder and universal tokens to extract causal signals from TabPFN's data embeddings and decode them into adjacency matrices. Our approach combines insights from tabular prompt-tuning methods \citep{tunetables2024} as we introduce universal tokens that are tuned to aggregate the causal information via our decoder for a given dataset in-context, in contrast to dataset-specific tuning, while following the neural causal discovery framework of \citet{lorch2022avici} in terms of problem formulation.

    \paragraph{Contributions.}
    We make three primary contributions: (1) introducing a novel research direction probing tabular foundation models for implicit causal knowledge, (2) developing a causal discovery framework that depends on foundation model representations, and (3) demonstrating that causal information is concentrated in the pre-trained TabPFN's middle layers.

\section{Related work and background}
    
    \textbf{Causal discovery} Traditional methods exploit observational and interventional data to extract causal structure following different approaches, including constraint-based \citep{spirtes2000pc}, score-based \citep{chickering2002ges}, continuous optimization \citep{zheng2018notears}, and interventional methods \citep{hauser2012characterization, wang2017permutation, brouillard2020differentiable}. While many of these approaches offer strong guarantees, they often make overly strict assumptions about the data generating process, including the nature of the noise signal \citep{shimizu2006lingam}, and face exponential computational complexity as feature sizes increase, leading to an increasing number of statistical tests \citep{recursive2025CD}. In recent years, neural causal discovery methods \citep{lorch2022avici, ke2023csiva, dhir2025a} address these limitations in an end-to-end manner without imposing the same limiting assumptions about data in their design, leveraging transformers trained on synthetic datasets for efficient single-pass causal graph prediction that is scalable to increasing feature sizes. Unlike these methods that are trained specifically for the causal discovery task, we exploit existing foundation models that may encode causal knowledge through their pre-training on predictive tasks.

    \textbf{Prior-Data Fitted Networks and causality} Prior-Data Fitted Networks (PFNs) are models pre-trained on synthetic datasets to perform a variety of predictive tasks. TabPFN \citep{hollmann2022tabpfn, hollmann2025tabpfnv2} is pre-trained on datasets generated from structural causal models (SCMs) and has achieved state-of-the-art performance for classification and regression. Recent extensions of PFNs to causal inference tasks, including causal fairness and treatment effect prediction \citep{robertson2024fairpfn, robertson2025dopfn, ma2025foundation}, demonstrate that PFNs can be trained to encode causal representations, which motivates us to investigate whether TabPFNv2, due to its causal prior, encodes causal knowledge.
    
    \textbf{AVICI} \citet{lorch2022avici} introduced a causal discovery framework that amortizes causal structure learning. Similar to TabPFN, it employs a dual-attention encoder to process the data, whose embeddings are aggregated into feature-wise representations, each corresponding to a graph node. These representations are used to predict the adjacency matrix in a pairwise manner. We adopt the same approach for adjacency prediction (Section \ref{subsec:architecture}) and loss function (Section \ref{subsec:loss}). To ensure a standardized comparison, we also use the same synthetic data-generating pipeline (Section \ref{subsec:data_generation}). 
        
   \section{Methodology}
    \label{methodology}
        
    \subsection{Architecture design}
    \label{subsec:architecture}

\begin{figure}[t]
        \centering
        \includegraphics[width=\columnwidth]{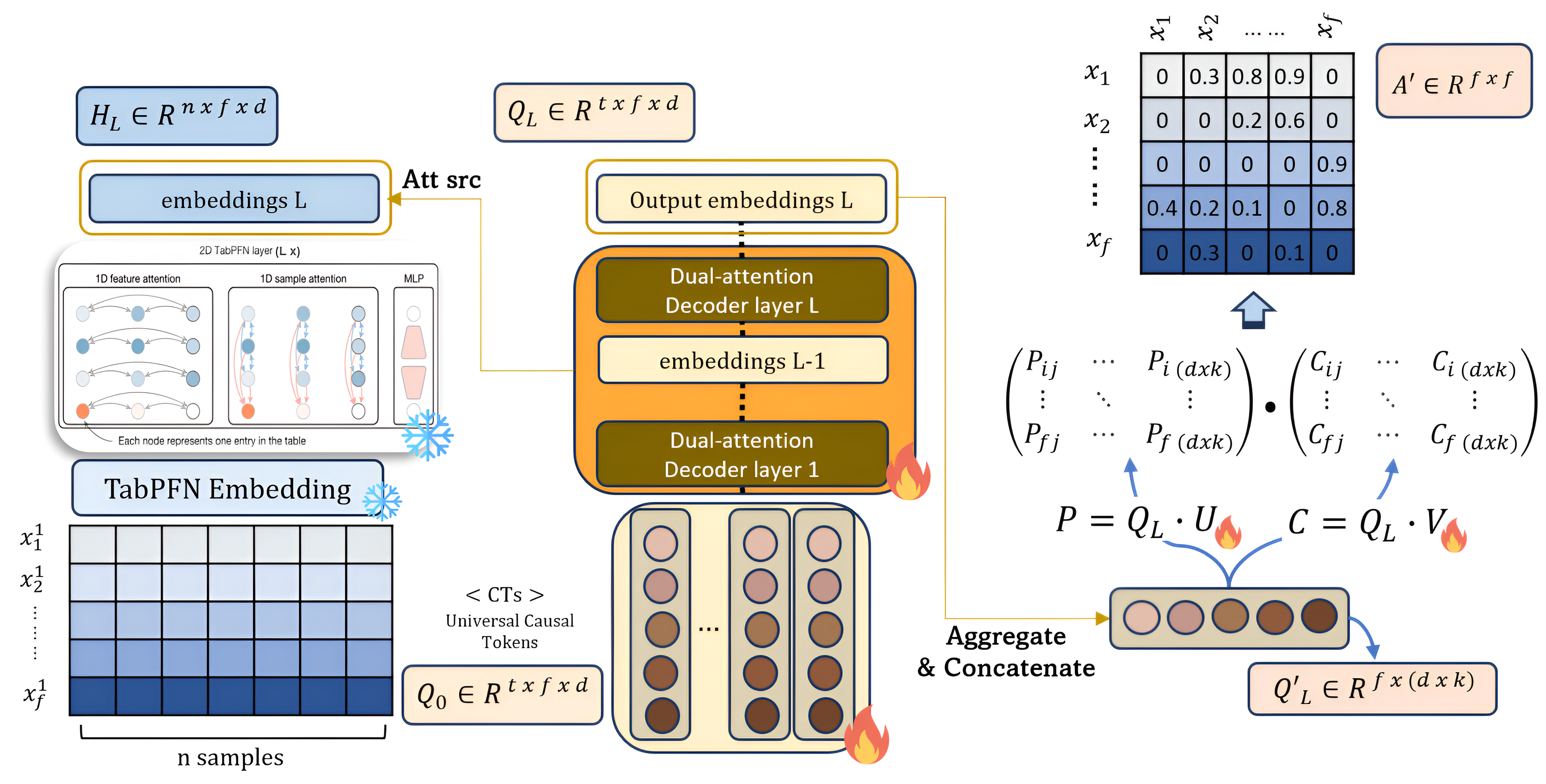}
        \caption{Overall architecture of our approach, where the data embeddings from the frozen TabPFN (left) are attended to in the decoder (middle) to extract aggregated feature-representations for the adjacency matrix prediction (right)}
        \label{fig:architecture}
    \end{figure}

Our proposed architecture is illustrated in Figure~\ref{fig:architecture}. We use TabPFNv2's classification backbone encoder with frozen weights and introduce $t$ learnable universal causal tokens $Q_0 \in \mathbb{R}^{t \times f \times d}$, which are prompt-tuned for causal discovery within a learnable dual-attention decoder that mirrors the encoder architecture but differs in its attention source. Our architecture proceeds as follows for a dataset with $f$ features and $n$ samples: (1) We obtain cell-wise $d$-dimensional representations via TabPFN's frozen embedding layer to produce data embeddings $H_0 \in \mathbb{R}^{n \times f \times d}$. These embeddings pass through the first four layers ($L = 4$) of TabPFN's dual-attention encoder, yielding data tokens $H_L$. (2) In our learnable decoder, causal tokens attend to these data tokens at each layer, summarizing causal information and producing output tokens $R_L \in \mathbb{R}^{t \times f \times d}$. (3) We aggregate the decoder outputs across the $t$ dimension into $k$ tokens ($k < t$), which are concatenated along the representational dimension to form expressive feature-wise representations $\in \mathbb{R}^{f \times k}$. (4) These representations are linearly projected via learnable matrices $U, V \in \mathbb{R}^{k \times k}$ into parent and child embeddings, and a dot product is applied to predict adjacency entries for each parent--child pair. (5) Finally, we apply a sigmoid activation to obtain predicted edge probabilities. Further details are provided in Appendix~\ref{app:architecture}.

    \subsection{Objective function}
    \label{subsec:loss}

    The objective is to amortize causal structure inference by maximizing the log-likelihood of the ground-truth adjacency matrix, approximated via binary cross-entropy over the edges represented as binary entries in the matrix. Training also promotes acyclicity of the predicted adjacency matrix, estimated by its spectral radius following \cite{lee2019acyclicity}, through constrained optimization. Further details are provided in Appendix \ref{app:loss_function}.

    \subsection{Data generation}\label{subsec:data_generation}

        Synthetic data are generated by sampling a directed acyclic graph (DAG) and then drawing samples from it under causal sufficiency. For each dataset, all parent--child mechanisms are defined using either a linear or a random Fourier feature (RFF) function. Each variable is sampled conditionally on its parents according to the chosen functional type. For details about graph structures, mechanisms, and noise types, see Appendix~\ref{app:data_generation}.

\section{Experiments}
    \subsection{Training and evaluation details}
    \label{sec:training}

       The model was trained for 100,000 optimization steps with a batch size of 32 datasets and their corresponding DAGs, using the AdamW optimizer \citep{loshchilov-iclr18a} with cosine annealing \citep{loshchilov-iclr17a} (initial learning rate = 5e-4). For each training batch, the feature dimensionality of datasets was randomly sampled between 4 and 20, proportional to the total number of features. Each dataset was generated according to one of two sampling schemes: (i) with probability 0.75, consisting of 100 observational and 100 interventional samples; or (ii) with probability 0.25, consisting solely of 200 observational samples. The number of causal tokens used is $t=30$, aggregated into $k=4$ tokens. Our model has 6M parameters,  3.6M of which are learnable.

        We evaluated on 500 datasets with their corresponding DAGs, spanning feature sizes of 5, 7, 10, 15, and 20. Each dataset contains 300 observational and 300 interventional samples. Our performance is compared against the pre-trained AVICI (scm-v0) model and statistical baselines for causal discovery, namely GIES \citep{hauser2012characterization}, IGSP \citep{wang2017permutation}, and DCDI \citep{brouillard2020differentiable}. Due to computational constraints, DCDI was evaluated on only 50 datasets. We report the ROC AUC and AP scores given the probabilistic binary classification nature of our edge predictions.
    
    \subsection{Results} 
    \label{sec:results}

        Our experiments are geared to answer the following research questions, with the key message that our framework can extract causal information from TabPFN and outperforms statistical baselines in the downstream task of causal discovery.

\begin{figure}[t]
            \centering
            \includegraphics[width=\textwidth]{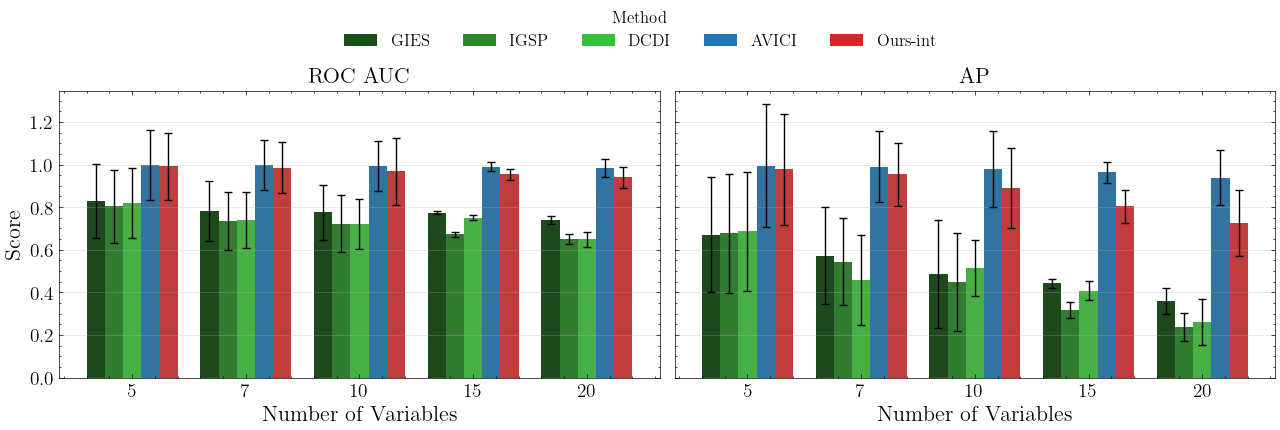}
            \caption{While we outperform statistical baselines (in green), we perform closely to AVICI on ROC AUC (left) in a stable manner, yet witness an increasing degradation in AP at scale (right).}
            \label{fig:roc_ap_bar}
        \end{figure}

    \paragraph{RQ1: Can we extract causal information from TabPFN embeddings?} First, we empirically show that our approach extracts causal information by achieving ROC AUC scores close to AVICI (Figure \ref{fig:roc_ap_bar}). While we outperform statistical methods on ROC AUC and AP scores, we observe that the AP scores decline at increasing feature sizes for our approach and the statistical baselines, suggesting that while TabPFN encodes causal structure, it struggles to distinguish the correct causal relations as the number of possible edges increases. Appendix \ref{app:ablations} further analyzes how performance varies across graph structures, likely due to TabPFN's pre-training on small and sparse graphs for predictive tasks.
        
\begin{figure}[H]
        \centering
        \begin{minipage}{0.48\textwidth}
            \centering
            \includegraphics[width=\textwidth]{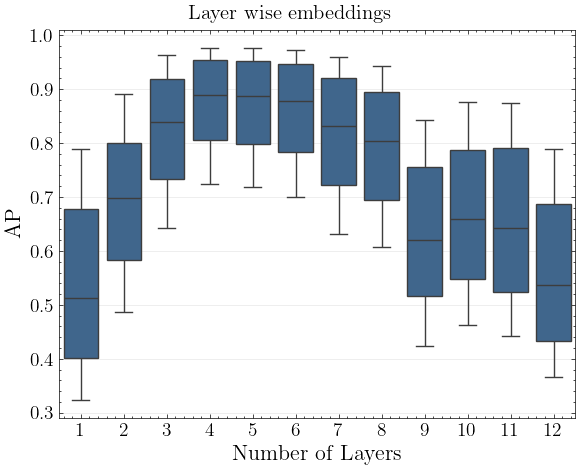}
        \end{minipage}
        \hfill
        \begin{minipage}{0.48\textwidth}
            \centering
            \includegraphics[width=\textwidth]{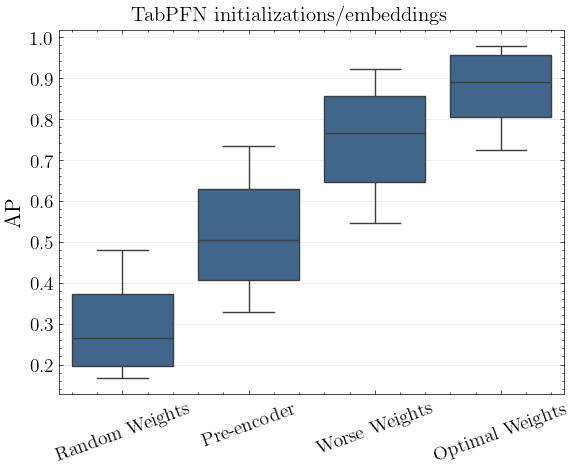}
        \end{minipage}
        \caption{AP scores of our approach trained using different encoder/decoder layers (left) and different initializations/embeddings (right), showing that the middle layers and that embeddings from TabPFN's encoder of optimal weights encode better causal information.}
        \label{fig:layer_encoder_abl}
    \end{figure}

    \paragraph{RQ2: How does causal information propagate through TabPFN?} Figure \ref{fig:layer_encoder_abl} shows that relying on data embeddings from the middle layers of TabPFN, namely layers 4-6, results in better causal understanding in comparison to the first and last layers. This aligns with interpretations about how models can define functional understanding in the middle layers, with layers towards the end being more adapted to the downstream tasks \citep{sia2024where,earlyexit2025}.
    
    \paragraph{RQ3: How important is the encoder?} We investigate whether TabPFN’s encoder captures causal structure or if our performance is mainly attributed to the decoder. Figure \ref{fig:layer_encoder_abl} compares embeddings from four variants: the official pre-trained encoder (Optimal Weights), a randomly initialized encoder (Random Weights), embeddings before the encoder (Pre-encoder), and a fine-tuned version trained to perform worse on classification tasks (Worse Weights). Our results show that using weights and data embeddings from the pre-trained encoder improves performance in causal discovery, while degrading predictive performance (Worse Weights) also diminishes causal accuracy, suggesting that TabPFN’s pre-training encodes feature interactions aligned with underlying causal relationships. Appendix \ref{app:ablations} further shows that the decoder architecture has a comparably strong impact.
    
    \section{Conclusion and future work}
    \label{sec:conclusion}
    
        We show that TabPFN's embeddings contain causal information and that our adaptor framework outperforms traditional causal discovery algorithms when causal information is extracted from mid-range layers. This further promotes leveraging pre-trained tabular models for extracting causal structures, improving the interpretability of these models, and aiding in scientific discovery. Our framework can further help understand how tabular foundation models reason about different datasets, and provides a way to repurpose tabular foundation models for different downstream tasks.

\section*{Acknowledgments}
This research was partially supported by the Deutsche Forschungsgemeinschaft (DFG, German Research Foundation) under
SFB 1597 (SmallData), grant number 499552394. The authors acknowledge support by the state of Baden-Württemberg through bwHPC and the German Research Foundation (DFG) through grant INST 35/1597-1 FUGG. Frank Hutter acknowledges financial support by the Hector Foundation. Omar Swelam acknowledges support by
the Konrad Zuse School of Excellence in Learning and Intelligent Systems (ELIZA) through the DAAD programme Konrad Zuse
Schools of Excellence in Artificial Intelligence, sponsored by the Federal Ministry of Education and Research. The authors acknowledge support from ELLIS and ELIZA. Funded by the European Union. Views and opinions expressed are however those of the author(s) only and do not necessarily reflect those of the European Union.

\bibliographystyle{icml2024}
\bibliography{references}

\begin{thebibliography}{26}
\providecommand{\natexlab}[1]{#1}
\providecommand{\url}[1]{\texttt{#1}}
\expandafter\ifx\csname urlstyle\endcsname\relax
  \providecommand{\doi}[1]{doi: #1}\else
  \providecommand{\doi}{doi: \begingroup \urlstyle{rm}\Url}\fi

\bibitem[Barab{\'a}si \& Albert(1999)Barab{\'a}si and Albert]{barabasi1999emergence}
Barab{\'a}si, A.-L. and Albert, R.
\newblock Emergence of scaling in random networks.
\newblock \emph{science}, 286\penalty0 (5439):\penalty0 509--512, 1999.

\bibitem[Brouillard et~al.(2020)Brouillard, Lachapelle, Lacoste, Lacoste-Julien, and Drouin]{brouillard2020differentiable}
Brouillard, P., Lachapelle, S., Lacoste, A., Lacoste-Julien, S., and Drouin, A.
\newblock Differentiable causal discovery from interventional data.
\newblock In \emph{Advances in Neural Information Processing Systems}, volume~33, pp.\  21865--21877, 2020.

\bibitem[Chickering(2002)]{chickering2002ges}
Chickering, D.~M.
\newblock Optimal structure identification with greedy search.
\newblock \emph{Journal of Machine Learning Research}, 2002.

\bibitem[Dhir et~al.(2025)Dhir, Ashman, Requeima, and van~der Wilk]{dhir2025a}
Dhir, A., Ashman, M., Requeima, J., and van~der Wilk, M.
\newblock A meta-learning approach to bayesian causal discovery.
\newblock In \emph{The Thirteenth International Conference on Learning Representations}, 2025.

\bibitem[Feuer et~al.(2024)Feuer, Schirrmeister, Cherepanova, Hegde, Hutter, Goldblum, Cohen, and White]{tunetables2024}
Feuer, B., Schirrmeister, R.~T., Cherepanova, V., Hegde, C., Hutter, F., Goldblum, M., Cohen, N., and White, C.
\newblock Tunetables: Context optimization for scalable prior-data fitted networks.
\newblock \emph{Advances in Neural Information Processing Systems}, 37:\penalty0 83430--83464, 2024.

\bibitem[Gilbert(1961)]{gilbert1961random}
Gilbert, E.~N.
\newblock Random plane networks.
\newblock \emph{Journal of the society for industrial and applied mathematics}, 9\penalty0 (4):\penalty0 533--543, 1961.

\bibitem[Hauser \& B{\"u}hlmann(2012)Hauser and B{\"u}hlmann]{hauser2012characterization}
Hauser, A. and B{\"u}hlmann, P.
\newblock Characterization and greedy learning of interventional markov equivalence classes of directed acyclic graphs.
\newblock \emph{The Journal of Machine Learning Research}, 13\penalty0 (1):\penalty0 2409--2464, 2012.

\bibitem[Holland et~al.(1983)Holland, Laskey, and Leinhardt]{holland1983stochastic}
Holland, P.~W., Laskey, K.~B., and Leinhardt, S.
\newblock Stochastic blockmodels: First steps.
\newblock \emph{Social networks}, 5\penalty0 (2):\penalty0 109--137, 1983.

\bibitem[Hollmann et~al.(2022)Hollmann, M{\"u}ller, Eggensperger, and Hutter]{hollmann2022tabpfn}
Hollmann, N., M{\"u}ller, S., Eggensperger, K., and Hutter, F.
\newblock Tabpfn: A transformer that solves small tabular classification problems in a second.
\newblock \emph{arXiv preprint arXiv:2207.01848}, 2022.

\bibitem[Hollmann et~al.(2025)Hollmann, M{\"u}ller, Purucker, Krishnakumar, K{\"o}rfer, Hoo, Schirrmeister, and Hutter]{hollmann2025tabpfnv2}
Hollmann, N., M{\"u}ller, S., Purucker, L., Krishnakumar, A., K{\"o}rfer, M., Hoo, S.~B., Schirrmeister, R.~T., and Hutter, F.
\newblock Accurate predictions on small data with a tabular foundation model.
\newblock \emph{Nature}, 01 2025.
\newblock \doi{10.1038/s41586-024-08328-6}.

\bibitem[Ke et~al.(2023)Ke, Chiappa, Wang, Bornschein, Goyal, Rey, Weber, Botvinick, Mozer, and Rezende]{ke2023csiva}
Ke, N.~R., Chiappa, S., Wang, J.~X., Bornschein, J., Goyal, A., Rey, M., Weber, T., Botvinick, M., Mozer, M.~C., and Rezende, D.~J.
\newblock Learning to induce causal structure.
\newblock In \emph{International Conference on Learning Representations}, 2023.

\bibitem[K{\"u}ken et~al.(2025)K{\"u}ken, Purucker, and Hutter]{earlyexit2025}
K{\"u}ken, J., Purucker, L., and Hutter, F.
\newblock Early stopping tabular in-context learning.
\newblock \emph{arXiv preprint arXiv:2506.21387}, 2025.

\bibitem[Lee et~al.(2019)Lee, Danieletto, Miotto, Cherng, and Dudley]{lee2019acyclicity}
Lee, H.-C., Danieletto, M., Miotto, R., Cherng, S.~T., and Dudley, J.~T.
\newblock Scaling structural learning with {NO-BEARS} to infer causal transcriptome networks.
\newblock In \emph{Pacific Symposium on Biocomputing 2020}, pp.\  391--402. World Scientific, 2019.

\bibitem[Lorch et~al.(2022)Lorch, Sussex, Rothfuss, Krause, and Sch{\"o}lkopf]{lorch2022avici}
Lorch, L., Sussex, S., Rothfuss, J., Krause, A., and Sch{\"o}lkopf, B.
\newblock Amortized inference for causal structure learning.
\newblock \emph{Advances in Neural Information Processing Systems}, 35:\penalty0 13104--13118, 2022.

\bibitem[Loshchilov \& Hutter(2017)Loshchilov and Hutter]{loshchilov-iclr17a}
Loshchilov, I. and Hutter, F.
\newblock {SGDR}: Stochastic gradient descent with warm restarts.
\newblock In \emph{International Conference on Learning Representations}, 2017.

\bibitem[Loshchilov \& Hutter(2019)Loshchilov and Hutter]{loshchilov-iclr18a}
Loshchilov, I. and Hutter, F.
\newblock Decoupled weight decay regularization.
\newblock In \emph{International Conference on Learning Representations}, 2019.

\bibitem[Ma et~al.(2025)Ma, Frauen, Javurek, and Feuerriegel]{ma2025foundation}
Ma, Y., Frauen, D., Javurek, E., and Feuerriegel, S.
\newblock Foundation models for causal inference via prior-data fitted networks.
\newblock \emph{arXiv preprint arXiv:2506.10914}, 2025.

\bibitem[Mokhtarian et~al.(2025)Mokhtarian, Elahi, Akbari, and Kiyavash]{recursive2025CD}
Mokhtarian, E., Elahi, S., Akbari, S., and Kiyavash, N.
\newblock Recursive causal discovery.
\newblock \emph{Journal of Machine Learning Research}, 26\penalty0 (61):\penalty0 1--65, 2025.

\bibitem[Robertson et~al.(2024)Robertson, Hollmann, Awad, and Hutter]{robertson2024fairpfn}
Robertson, J., Hollmann, N., Awad, N., and Hutter, F.
\newblock Fairpfn: Transformers can do counterfactual fairness.
\newblock \emph{arXiv preprint arXiv:2407.05732}, 2024.

\bibitem[Robertson et~al.(2025)Robertson, Reuter, Guo, Hollmann, Hutter, and Sch{\"o}lkopf]{robertson2025dopfn}
Robertson, J., Reuter, A., Guo, S., Hollmann, N., Hutter, F., and Sch{\"o}lkopf, B.
\newblock Do-{PFN}: In-context learning for causal effect estimation.
\newblock \emph{arXiv preprint arXiv:2506.06039}, 2025.

\bibitem[Shimizu et~al.(2006)Shimizu, Hoyer, Hyv{\"a}rinen, Kerminen, and Jordan]{shimizu2006lingam}
Shimizu, S., Hoyer, P.~O., Hyv{\"a}rinen, A., Kerminen, A., and Jordan, M.
\newblock A linear non-gaussian acyclic model for causal discovery.
\newblock \emph{Journal of Machine Learning Research}, 7\penalty0 (10), 2006.

\bibitem[Sia et~al.(2024)Sia, Mueller, and Duh]{sia2024where}
Sia, S., Mueller, D., and Duh, K.
\newblock Where does in-context learning happen in large language models?
\newblock \emph{Advances in Neural Information Processing Systems}, 37:\penalty0 32761--32786, 2024.

\bibitem[Spirtes et~al.(2000)Spirtes, Glymour, and Scheines]{spirtes2000pc}
Spirtes, P., Glymour, C., and Scheines, R.
\newblock Causation, prediction, and search.
\newblock \emph{MIT Press}, 2000.

\bibitem[Wang et~al.(2017)Wang, Solus, Yang, and Uhler]{wang2017permutation}
Wang, Y., Solus, L., Yang, K., and Uhler, C.
\newblock Permutation-based causal inference algorithms with interventions.
\newblock \emph{Advances in Neural Information Processing Systems}, 30, 2017.

\bibitem[Watts \& Strogatz(1998)Watts and Strogatz]{watts1998collective}
Watts, D.~J. and Strogatz, S.~H.
\newblock Collective dynamics of ‘small-world’networks.
\newblock \emph{nature}, 393\penalty0 (6684):\penalty0 440--442, 1998.

\bibitem[Zheng et~al.(2018)Zheng, Aragam, Ravikumar, and Xing]{zheng2018notears}
Zheng, X., Aragam, B., Ravikumar, P.~K., and Xing, E.~P.
\newblock Dags with no tears: Continuous optimization for structure learning.
\newblock \emph{Advances in neural information processing systems}, 31, 2018.

\end{thebibliography}

\newpage
\appendix

\section{Architecture}\label{app:architecture}

Our architecture integrates three main components that interact sequentially to extract and model causal structure from tabular data. First, TabPFN's pre-trained embedding layer and encoder transform each value in a dataset into a contextualized embedding that captures relationships across samples and features. These embeddings are then processed by a decoder equipped with learnable causal tokens, which attend to the encoder outputs to identify potential causal directions. Finally, a DAG prediction module aggregates the causal token representations and produces pairwise edge probabilities, forming the estimated causal graph.

\subsection{TabPFN encoder architecture and input preprocessing}
\label{app:encoder}
    Our encoder leverages the pre-trained TabPFNv2 \citep{hollmann2025tabpfnv2} architecture, which now produces a dedicated representation for each feature value in each row in the input dataset.
    
    Specifically, we embed the input data matrix $X \in \mathbb{R}^{n\times f}$ into a tensor $\mathcal{H} \in \mathbb{R}^{n\times f\times d}$, where $d$ is the embedding dimension. This is accomplished by passing each entry $X_{ij}$ through a TabPFN-learned projection layer that maps scalar cell values into $d$-dimensional vectors where $d=192$.

     To differentiate between interventional and observational samples, each feature in each sample is assigned a binary label indicating whether it is interventional or observational. This makes our data input shape different from the original TabPFN input shape, which is $(n, f)$, to be $(n, f, 2)$ so we have two values for each feature in a datapoint, one indicating its real value and another indicating its interventional value. To get the per-feature representation, each 2 feature values are grouped into a single representation, so the input shape of $(n, f, 2)$ is then encoded by TabPFN's initial projection layer into $(n, f, d)$ as needed. It is worth noting that given the causal sufficiency assumption, each feature corresponds to a node in the DAG generating the dataset.

    Following the projection encoding layer, TabPFN's main encoder architecture uses a transformer encoder with a dual-attention mechanism that processes information along both the sample and feature dimensions via self-attention. After $L$ Transformer layers, each sample token's state encodes information from other samples as well as from different features. The encoder's final output is $\mathcal{H}_{out} \in \mathbb{R}^{n\times f\times d}$. While TabPFN's original architecture has 12 layers ($L = 12$), we use $L=4$ for our experiments, given the findings shown in Figure \ref{fig:layer_encoder_abl}.

\subsection{Causal tokens and decoder}
\label{app:tokens_decoder}

Our main approach is dependent on having universal tokens that are tuned to capture the causal effect direction, which is inspired by TuneTables ~\citep{tunetables2024}. We initialize the causal tokens as learnable embeddings of the shape $(20, t \times d)$ where $t=30$ and 20 is set as the maximum number of features we considered in our setup. For datasets with fewer than $f$ features, we use only the first $f$ embeddings, which are then reshaped from $(f, t \times d)$ to $(t, f, d)$.

The learnable decoder performs cross-attention with the data tokens $\mathcal{H}_L \in \mathbb{R}^{n \times f \times d}$, obtained from layer $L$ of TabPFN's frozen encoder, serving as the attention \emph{keys} and \emph{values}, while the causal tokens act as the \emph{queries}. This design allows the causal tokens to selectively attend to relevant data representations and progressively integrate causal information across layers.

Structurally, the decoder mirrors the encoder’s dual-attention architecture, where each layer contains two alternating multi-head attention blocks applied \emph{across features} and \emph{across samples}, interleaved with feed-forward sublayers. The key difference lies in the attention direction: instead of self-attending over data tokens, the decoder performs cross-attention from causal tokens to data tokens in both attention types. The number of layers and representational dimensionality are identical to those of the encoder to maintain architectural symmetry and stable information flow.

In our framework, the decoder is trained to encapsulate the causal relationships encoded in the data tokens into the causal tokens $\mathcal{Q}_L \in \mathbb{R}^{t \times f \times d}$. These causal tokens thus serve as compact, learnable summaries that aggregate structural dependencies across features, which are later decoded into the predicted adjacency matrix.

\subsection{DAG prediction}
\label{app:aggregate_decode}

To improve computational efficiency while preserving statistical richness, we aggregate the information across the $t$ causal tokens into $k = 4$ representative tokens by applying four statistical operations, namely \textit{max}, \textit{min}, \textit{mean}, and \textit{std}. The resulting $k$ tokens are concatenated along the representational dimension to form a matrix of shape $(f, k \times d)$ containing per-feature representations. Following~\cite{lorch2022avici}, for each feature $i$ ($i = 1, 2, \ldots, f$), we apply linear projections to obtain a parent representation $V_i \in \mathbb{R}^{k \times d}$ and a child representation $U_i \in \mathbb{R}^{k \times d}$. The pairwise adjacency score between features $i$ and $j$ is then computed as the dot product of their respective parent and child representations, followed by a sigmoid activation to yield the edge probability.

\section{Loss function}\label{app:loss_function}

\textbf{Binary cross-entropy loss:} The binary cross-entropy loss formulation is:
\begin{equation}
    \mathcal{L} = - \frac{1}{f(f-1)} \sum_{i \neq j} \big[ A_{ij} \log \hat{A}_{ij} + (1 - A_{ij}) \log (1 - \hat{A}_{ij}) \big],
\end{equation}
where $f$ is the number of features and $A_{ij}$ is the ground-truth adjacency matrix. The loss is weighted based on the number of edges in the graph to balance the learning of the model across different numbers of features.

\textbf{Acyclicity constraint:} For constraining the predicted adjacency matrix to be acyclic, the power iteration method is used to estimate the largest eigenvalue of the adjacency matrix to maintain numerical stability. Such constrained optimization follows a dual formulation for our model's learnable parameters through an augmented Lagrangian approach where dual variables track constraint violations, and the penalty weights are dynamically adjusted during training.

\section{Data generation}\label{app:data_generation}

When sampling a dataset, we first sample the graph structure used to construct the directed acyclic graph (DAG). To generate the dataset values, one data generating function type, either linear or random Fourier feature (RFF), is then randomly selected and applied to define the functional mechanisms of all nodes. 

\subsection{Graph structures}\label{app:graph_structures}

\begin{figure}[H]
    \centering
    \includegraphics[width=\textwidth]{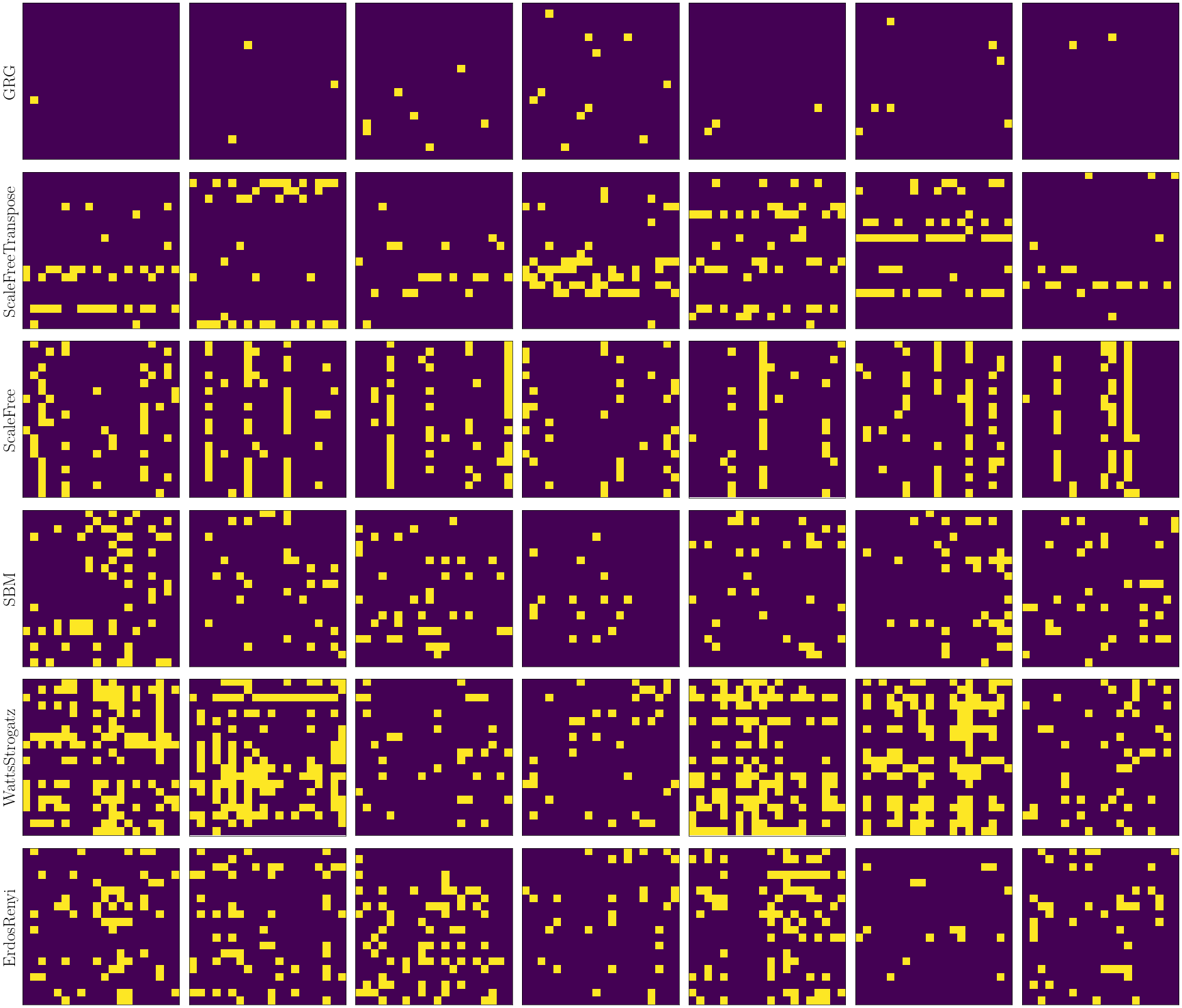}
    \caption{Samples of different graphs sampled from the graph structures. Each plot represents the adjacency matrix of a sampled DAG, where the yellow dots represent the edges.}
    \label{fig:graph_structures}
\end{figure}

The DAG generation process employs multiple graph generation mechanisms to ensure diverse causal topologies during training, mainly using five distinct graph generation processes. Figure \ref{fig:graph_structures} shows samples from the different graph structures listed below:

\paragraph{Erdős-Rényi Random Graphs:} Each edge is sampled independently with a fixed probability, serving as a baseline topology.

\paragraph{Scale-Free Graphs:} Incoming or outgoing edges of nodes are added to the previous node with probability proportional to $\deg(i)^\alpha$ \citep{barabasi1999emergence}. This creates graphs with heavy-tailed degree distributions commonly observed in biological and social networks.

\paragraph{Watts-Strogatz Small-World Networks:} These are $k$-dimensional lattices where edges are rewired globally to random nodes \citep{watts1998collective}.

\paragraph{Stochastic Block Model:} This model generates graphs with community structure by first partitioning nodes into random blocks, then setting inter-block edge probabilities lower than intra-block probabilities, capturing hierarchical structures found in complex systems \citep{holland1983stochastic}.

\paragraph{Geometric Random Graphs:} Nodes are randomly placed in a unit square, and edges are formed based on two-dimensional Euclidean distance below a threshold \citep{gilbert1961random}.

\subsection{Data generating functions}\label{app:data_generating_functions}

The relationships within the SCM, corresponding to a dataset, are defined such that each causal variable $x_j$ is sampled given its parents $\xb_{\text{pa}(j)}$ as
\begin{align}
    x_j \gets f_j(\xb_{\text{pa}(j)}, \epsilon_j) = f_j(\xb_{\text{pa}(j)}) + h_j(\xb_{\text{pa}(j)})\varepsilon_j
\end{align}
where the noise $\varepsilon_j$ is additive. The data generating function $f_j$ can be either linear or a random Fourier feature (RFF) function as an approximation for a Gaussian Process. The noise term can be sampled either from a Gaussian, Laplace, or Cauchy distribution. The specification of the parameter space for the graph structures, data generating functions, and noise scales follows a similar setup to ~\cite{lorch2022avici}, to have a consistent framework to benchmark against.

\subsection{Interventions}\label{app:interventions}

The interventions in our generation process are single-variable interventions, where in each interventional data point, we intervene on a single variable.
Such intervention is only performed on a subset of half of the available features. The intervention values are sampled from a uniform distribution.

\subsection{Parameters of data generation}\label{app:data_generation_params}

\begin{table}[H]
    \centering
    \caption[Data generation parameters]{\looseness -1 A description of the data generation parameters used in our experiments. Graph structures are sampled with equal probability in all cases whenever specified within the domain of the distribution.}\label{tab:data_params}
    \vspace{15pt}
  \begin{adjustbox}{max width=\linewidth}
  \begin{threeparttable}
  \begin{tabular}{lll}
  \toprule
  & \textbf{Parameter} & \textbf{Values} \\
  \midrule
  \textbf{Graph} & & \\
  \midrule
  Erdős-Rényi & expected edges/node & $\in \{1, 2, 3\}$ \\
  Scale-free (in-degree) & edges/node & $\in \{1, 2, 3\}$ \\
  & attach.\ power $\alpha$ & $\in \{0.7, 1.0, 1.2, 1.5\}$ \\
  Scale-free (out-degree) & edges/node & $\in \{1, 2, 3\}$ \\
  & attach.\ power $\alpha$ & $\in \{0.7, 1.0, 1.2, 1.5\}$ \\
  Watts-Strogatz & lattice dim.\ $k$ & $\in \{2, 3\}$ \\
  & rewire prob. & $\in \{0.2, 0.4\}$ \\
  Stochastic Block Model & expected edges/node & $\in \{1, 2, 3\}$ \\
  & blocks & $\in \{2, 5, 10\}$ \\
  & damp.\ inter-block prob. & $\in \{0.1\}$ \\
  Geometric Random Graphs & radius & $\in \{0.08, 0.1, 0.15\}$ \\
  \midrule
  \\
  \textbf{Mechanism} & & \\
  \midrule
  Linear function & weights $\wb$ & $\sim \unifpm(0.25, 4)$ \\
  & bias $b$ & $\sim \unif(-3, 3)$ \\
  \midrule
  Random Fourier function & length scale $\ell$ & $\sim \unif(5, 12)$ \\
  & output scale $c$ & $\sim \unif(8, 22)$ \\
  & bias $b$ & $\sim \unif(-3, 3)$ \\
  \midrule
  \\
  \textbf{Noise} & & \\
  \midrule
  $\Ncal(0, \sigma^2)$ & $\sigma$ & $\sim \unif(0.2, 2)$ \\
  $\text{Laplace}(0, \sigma^2)$ & $\sigma^2(\xb_{\text{pa}_j})$ & $\sim p(h_\text{rff})$ \\
  $\text{Cauchy}(0, \sigma^2)$ & $\sigma^2(\xb_{\text{pa}_j})$ & $\sim p(h_\text{rff})$ \\
  \midrule
  \\
  \textbf{Interventions} & & \\
  \midrule
  Target nodes & selection & random 50\% of nodes \\
  Intervention values & $x_j$ & $\sim \unifpm(1, 5)$ \\
  \bottomrule
  \end{tabular}
  \vspace*{5pt}
  \begin{tablenotes}
  \footnotesize
  \vspace*{5pt}
  \item[] \hspace*{-10pt}Aliases:
  \begin{itemize}
  \item $\unifpm(a, b)$: uniform mixture of $\unif(a, b)$ and $\unif(-b, -a)$
  \item $p(h_\text{rff})$: distribution over heteroscedastic noise scale functions, induced by the squash function $h_\text{rff}(\xb) = \log(1+\exp(g_\text{rff}(\xb))$ and random Fourier feature functions $g_\text{rff}(\xb)$.
  \end{itemize}
  \end{tablenotes}
  \end{threeparttable}
  \end{adjustbox}
  \vspace{5pt}
  \end{table}

\section{Ablations}\label{app:ablations}
\subsection{Performance across decoder architectures}\label{app:decoder_architectures_abl}

\begin{figure}[h]
    \centering
    \includegraphics[width=\textwidth]{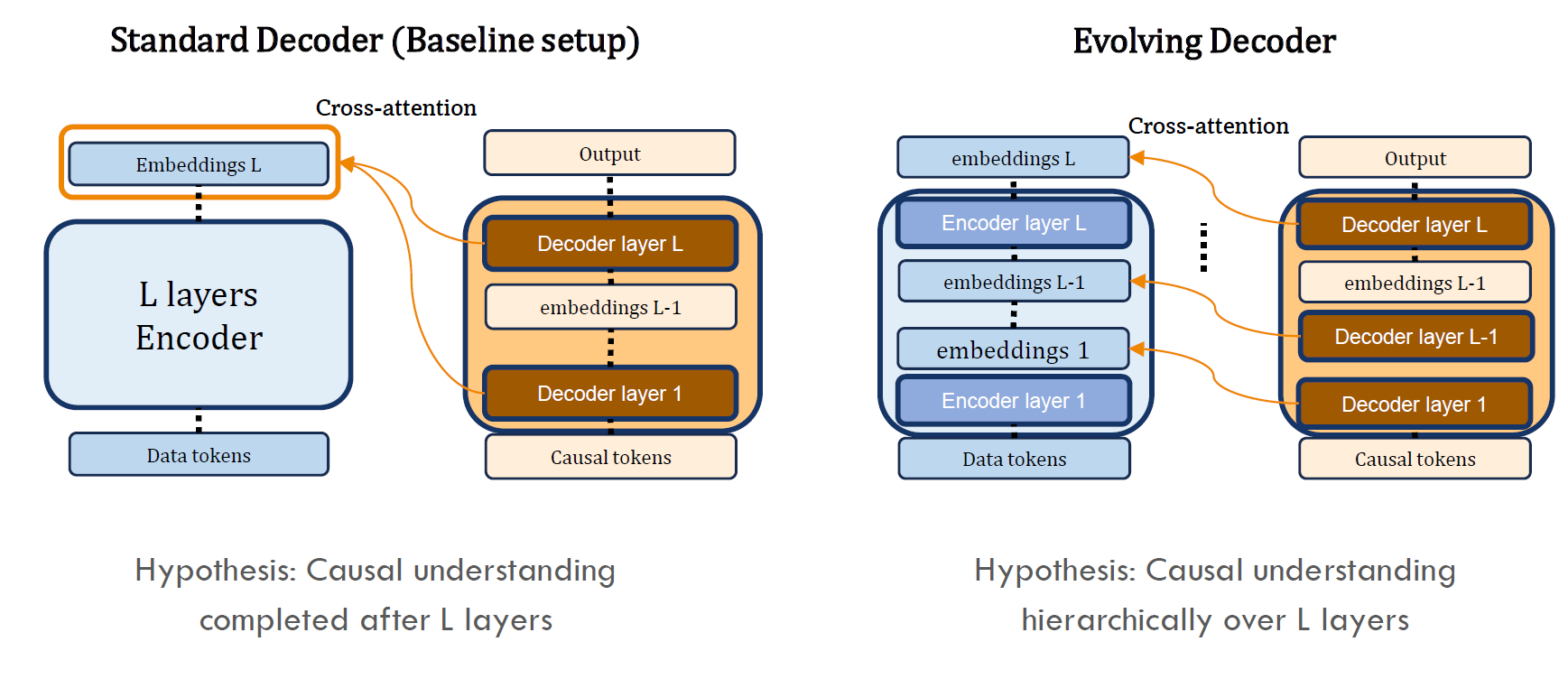}
    \caption{Different decoder architectures.}
    \label{fig:decoder_archs}
\end{figure}

While we have seen how the encoder choice affects our performance, we wanted to inspect how the decoder also plays a role. Therefore, we introduce two decoder setups as shown in Figure \ref{fig:decoder_archs} along with a "No Decoder" one:

\begin{itemize}
    \item \textbf{No Decoder:} We do not use a decoder and pass the causal tokens into TabPFN's encoder as query tokens as if they are $X_{test}$ data points. Thus, using the attention mechanism of TabPFN's encoder.
    \item \textbf{Standard Decoder:} We do cross attention with the attention source being data embeddings from the encoder's layer of choice, typically the 4th layer (our chosen approach).
    \item \textbf{Evolving Standard Decoder:} We do cross attention with the attention source in each decoder layer being data embeddings from the corresponding layer of the encoder, e.g., the 1st decoder layer attending to the output of the 1st encoder layer.
\end{itemize}

\begin{figure}[h]
    \centering
    \includegraphics[width=0.6\textwidth]{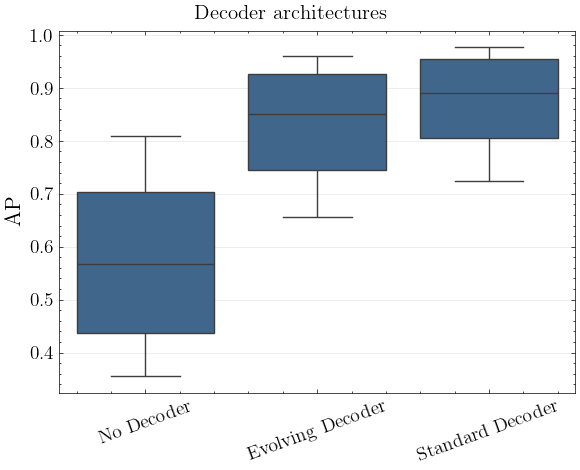}
    \caption{The presence of a decoder significantly improves the performance, showing that both the encoder's embeddings and the learnable decoder are essential for the causal information extraction.}
    \label{fig:decoder_abl}
\end{figure}

As shown in Figure \ref{fig:decoder_abl}, we have noticed how the performance of our approach varies across different decoder architectures, where having a learnable decoder plays a crucial role in the performance of our approach. The Standard Decoder setup witnesses the highest scores, while the No Decoder setup witnesses the lowest performance. This also indicates that the causal information can be considered completely evolved after the 4th layer of the encoder, not in a hierarchical manner.

\subsection{Performance across graph structures}\label{app:graph_structures_abl}

As shown before, we have noticed how the performance degrades across an increasing number of features in terms of AP scores. This further motivated us to analyze how such performance varies across different graph structures to understand what causes such degradation.

As shown below in Figure \ref{fig:ap_graph_int}, the scaling issue is less pronounced in graph structures that have central nodes with a higher density of edges or sparse setups (namely, Scale-Free and GRG, respectively), while the variance is more pronounced in dense graph structures with a relatively larger number of edges distributed across different nodes (namely, Erodos-Renyi, SBM, and Watts-Strogatz), as shown in the samples in Figure \ref{fig:graph_structures}. This can be attributed to the pre-training setup of TabPFN towards predictive tasks that aim at consolidating the information regarding a target variable, regardless of the rest of the feature interactions.
This is further supported by the performance of AVICI over the different graph structures. Although AVICI witnesses a lower performance on Watts-Strogatz, the performance is relatively stable across other structures compared to our approach.

\begin{figure}[h]
    \centering
    \includegraphics[width=\textwidth]{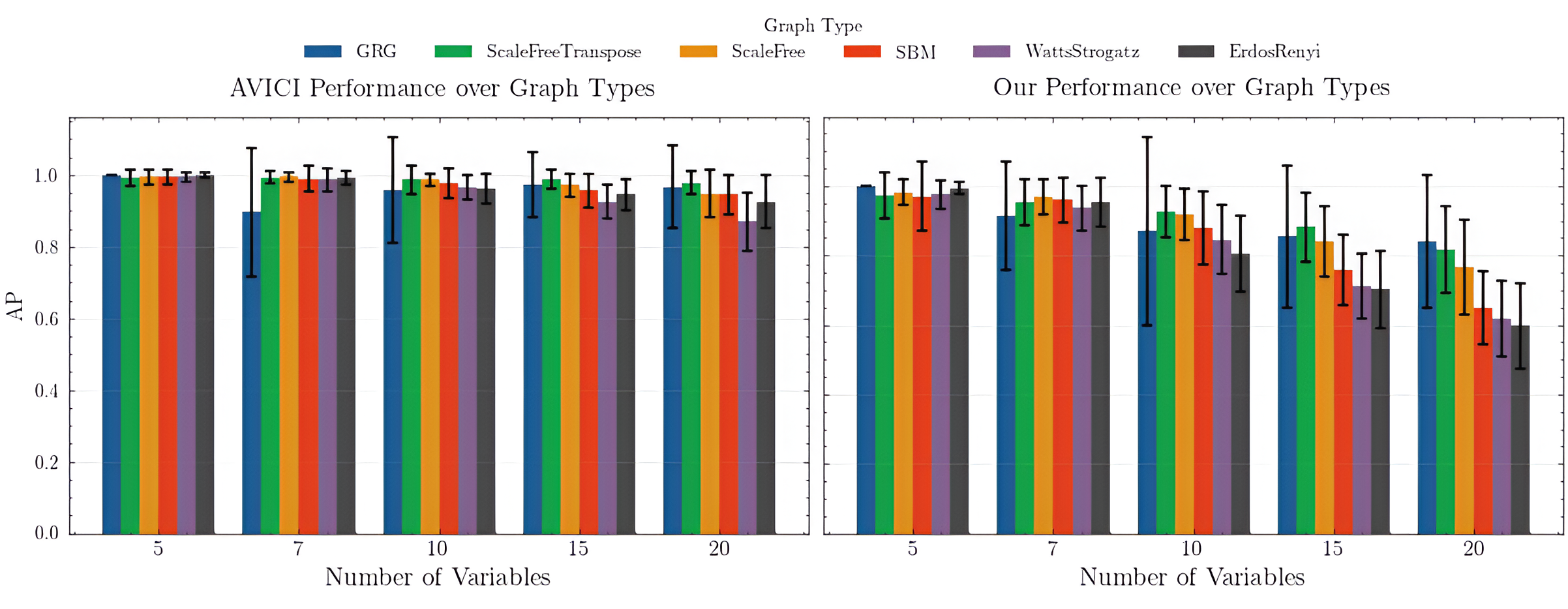}
    \caption{AP scores of our approach (right) compared to AVICI (left) for the different graph structures, showing how sensitive our approach is to the type of graph structures it witnesses at increasing scale.}
    \label{fig:ap_graph_int}
\end{figure}

\end{document}